\documentclass[twoside,english]{elsarticle}
\usepackage{ae,aecompl}
\usepackage[T1]{fontenc}
\usepackage[latin9]{inputenc}
\usepackage{color}
\usepackage{amstext}
\usepackage{graphicx}

\makeatletter

\newcommand{\lyxmathsym}[1]{\ifmmode\begingroup\def\b@ld{bold}
  \text{\ifx\math@version\b@ld\bfseries\fi#1}\endgroup\else#1\fi}

\journal{Pattern Recognition Letters}

\usepackage{lineno}

\definecolor{green}{RGB}{59, 211, 39}
\definecolor{red}{RGB}{192, 0, 0}
\definecolor{blue}{RGB}{0, 176, 240}
\definecolor{cyan}{RGB}{255, 0, 0}
\definecolor{magenta}{RGB}{0, 0, 255}
\definecolor{gray}{RGB}{211, 211, 211}
\definecolor{yellow}{RGB}{128, 128, 128}

\makeatother

\usepackage{babel}
\begin{document}

\title{Automated and Weighted Self-Organizing Time Maps}

\author[focal,rvt]{Peter Sarlin\corref{cor1}}

\cortext[cor1]{Corresponding author: RiskLab at IAMSR, Department of Information
Technologies, Åbo Akademi University. Postal address: Joukahaisenkatu
3-5, 20520 Turku, Finland. Email: psarlin@abo.fi. Tel: +358 2 215
4670.}

\address[focal]{Center of Excellence SAFE at Goethe University, Frankfurt, Germany}

\address[rvt]{RiskLab at IAMSR, Åbo Akademi University (Turku) and Arcada University
of Applied Sciences (Helsinki), Finland }
\begin{abstract}
This paper proposes schemes for automated and weighted Self-Organizing
Time Maps (SOTMs). The SOTM provides means for a visual approach to
evolutionary clustering, which aims at producing a sequence of clustering
solutions. This task we denote as visual dynamic clustering. The implication
of an automated SOTM is not only a data-driven parametrization of
the SOTM, but also the feature of adjusting the training to the characteristics
of the data at each time step. The aim of the weighted SOTM is to
improve learning from more trustworthy or important data with an instance-varying
weight. The schemes for automated and weighted SOTMs are illustrated
on two real-world datasets: (\emph{i}) country-level risk indicators
to measure the evolution of global imbalances, and (\emph{ii}) credit
applicant data to measure the evolution of firm-level credit risks.\end{abstract}
\begin{keyword}
Self-Organizing Time Map \sep weighting schemes\sep quality measures

\end{keyword}
\maketitle

\section{Introduction}

This paper proposes schemes for automated and weighted Self-Organizing
Time Maps (SOTMs). The SOTM \citep{Sarlin2012} is a recently introduced
adaptation of Kohonen's \citeyearpar{Kohonen1982} Self-Organizing
Map (SOM) that enables visual dynamic clustering. The SOTM provides
means for a visual approach to so-called evolutionary clustering (EC)
\citep{Chakrabartietal2006}, which aims at processing temporal data
by producing a sequence of clustering solutions. \citet{Chakrabartietal2006}
describe the aim of EC to be a balance between faithfulness of the
clustering to current data and comparability with the previous clustering
result, and relate it to four virtues: (\emph{i}) consistency (i.e.,
familiarity with the previous clustering), (\emph{ii}) noise removal
(i.e., a historical consistent clustering increases robustness), (\emph{iii})
smoothing (i.e., a smooth view of transitions), and (\emph{iv}) cluster
correspondence (i.e., relation to historical context). Recent research
has also provided other approaches to adaptive clustering, but with
different inspiration. Some examples constitute of kernel spectral
clustering with memory \citep{Langoneetal2013} and incremental $k$-means
\citep{ChakrabortyNagwani2011}. Beyond only performing EC, the SOTM
is a visual approach, as it provides means for a low-dimensional representation
of the high-dimensional data. The SOTM is essentially a sequence of
one-dimensional SOMs on data ordered in consequent time units and
represents both time and data topology on a two dimensional grid of
reference vectors. Hence, the topology preservation of the SOTM may
preserve data topology on the vertical direction and time topology
on the horizontal direction. 

The SOTM is applicable in a wide range of application areas with data
along three dimensions of a data cube: (\emph{i}) cross-section, (\emph{ii})
time, and (\emph{iii}) inputs. This is a common setting in many domains,
such as social sciences, finance and engineering, with data for several
entities, reference periods and explanatory variables. Since the introduction
of the approach, the SOTM has been combined with hierarchical clustering
methods to better reveal emerging, changing and disappearing clusters
\citep{SarlinYao2013} and shown to not be restricted to illustrating
changes over the time, but equally applicable for abstractions of
changes in cluster structures over any variable of ordinal, cardinal
or higher level of measurement \citep{Sarlin2013NC2}. The SOTM has
also been applied to a range of tasks, such as visual explorations
of cross-sectional changes in welfare and poverty \citep{Sarlin2012},
temporal customer segmentation \citep{Yaoetal2012b}, temporal customer
behavior analysis \citep{Yaoetal2013}, exploring and comparing green
\emph{vs}. non-green customer segmentations \citep{Holmbometal2013},
an abstraction of the ongoing global financial crisis \citep{SarlinPRL2013},
exploring structures in firm-level data \citep{SarlinYao2013} and
assessing risks related to fixed exchange rates \citep{SarlinYao2013}.
Motivated by the experience in these case studies, this paper aims
at enhancing the SOTM for it to be applicable in a wider setting.

In this paper, we propose schemes for automated and weighted SOTMs.
The first task relates to automating the parametrization with quantitative
quality measures. As the SOTM, like the standard SOM, has an aim related
to both quantification and topology-preservation accuracy, the parametrization
task commonly involves human judgment and manual experimentation.
Yet, the goodness measure proposed by \citet{KaskiLagus1996}, which
combines quantization and topographic errors, was proposed not only
to evaluate the goodness of a SOM, but also to compare a set of SOMs.
In this work, we adopt the Kaski-Lagus measure to the SOTM framework,
in order to achieve automated training of the sequence of SOMs, where
the parameters for each SOM (with a unique dataset) are derived using
the measure. The second task relates to modifying the priority or
weighting of data when training a SOTM. While a number of approaches
to weighting have been applied in the context of SOMs, the focus herein
is on learning that accounts for instance-varying importance. Following
the Weighted SOM (WSOM) \citep{SarlinWSOM2012}, we augment the neighborhood-function-based
weighting of the batch SOTM (e.g. Voronoi regions) with a user-specified,
instance-specific importance weight. This is a broader perspective
than incorporating a cost matrix into the objective function of a
classifier, which is most often the case in cost-sensitive learning,
as the aim of the weighted SOTM is  to improve learning from important
data with an instance-varying weight. To this end, it is not restricted
to classification tasks, but rather a feasible alternative for a cost-sensitive,
unsupervised version of visual dynamic clustering, as the weight could
also represent importance or trustworthiness of an instance for forming
clusters. The schemes for automated and weighted SOTMs are illustrated
on two real-world datasets: (\emph{i}) country-level risk indicators
to measure the evolution of global imbalances, and (\emph{ii}) credit
applicant data to measure the evolution of firm-level credit risks.

This paper is organized as follows. While Section 2 describes the
SOTM and its automation and weighting, Section 3 presents the experiments.
Section 4 concludes.

\section{Weighted and automated SOTMs}

This section presents the SOTM, and the automated and weighted training
schemes.

\subsection{The SOTM}

The SOTM \citep{Sarlin2012} is an approach for visual dynamic clustering.
Yet, it is nothing more than a sequence of one-dimensional SOMs with
a short-term memory. The SOMs are trained on data ordered in consequent
time units and short-term memory refers to initializations based upon
the values of the previous SOM. More formally, for time-coordinates
$t$ (where $t=1,2,\text{\ldots},T$), the SOTM performs a mapping
from the input data space $\Omega(t)$, with a probability density
function $p(x,t)$, onto a one-dimensional array $A(t)$ of output
units $m_{i}(t)$ (where $i=1,2,\text{\ldots},M$).

The orientation preservation is performed with the following initializations.
The first principal component of Principal Component Analysis is used
for initializing $A(t_{1})$ to set the orientation of the SOTM. To
preserve the orientation between consecutive patterns, the model uses
short-term memory. Thus, the orientation of the map is preserved by
initializing A($t_{2,3,...,T}$) with the reference vectors of $A(t-1)$.
The training at each $t_{1,2,\text{\ldots},T}$ is iterated for $s_{t}$
training steps, which points to the fact that the number of iterations
may vary over $t$ (as also proposed in \citep{Sarlin2012}). For
$A(t_{1,2,\text{\ldots},T})$, each data point $x_{j}(t)\in\Omega(t)$
(where $j=1,2,\lyxmathsym{\ldots},N(t)$) is compared to reference
vectors $m_{i}(t)\in A(t)$ and assigned to its BMU $m_{b}(t)$: 

\begin{equation}
\left\Vert x_{j}(t)-m_{b}(t,s_{t})\right\Vert =\min_{i}\left\Vert x_{j}(t)-m_{i}(t,s_{t})\right\Vert .\label{eq:SOTM_match}
\end{equation}
Then, each reference vector $m_{i}(t)$ is adjusted with a batch update: 

\begin{equation}
m_{i}(t,s_{t})=\frac{\sum_{j=1}^{N(t)}h_{ib(j)}(t,s_{t})x_{j}(t)}{\sum_{j=1}^{N(t)}h_{ib(j)}(t,s_{t})},\label{eq:SOTM_update}
\end{equation}
where index $j$ indicates the input data that belong to unit $b$
and the neighborhood function $h_{ib(j)}(t,s_{t})\in\left(0,1\right]$
is defined as a Gaussian function

\begin{equation}
h_{ib(j)}(t,s_{t})=\exp\left(-\frac{\left\Vert r_{b}(t)-r_{i}(t)\right\Vert ^{2}}{2\sigma^{2}(t,s_{t})}\right),\label{eq:SOTM_gaussian}
\end{equation}
where $\left\Vert r_{b}(t)-r_{i}(t)\right\Vert ^{2}$ is the squared
Euclidean distance between the coordinates of the reference vectors
$m_{b}(t)$ and $m_{i}(t)$ on the one-dimensional array, and $\sigma(t,s_{t})$
is the user-specified neighborhood parameter, which may vary over
time $t$ and training steps $s_{t}$. Relating to the originally
presented SOTM \citep{Sarlin2012}, the above introduced notation
is similar in nature but allows for training steps $s_{t}$ to be
larger than one and vary over time $t$, as well as allows for the
radius of the neighborhood $\sigma(t,s_{t})$ to change over time
$t$ and training steps $s_{t}$. This implies that the training at
each point in time may change as per the given needs.

\subsection{Automated SOTMs}

The notion of an automated SOTM relies on having an objective function
to minimize. Drawing upon the SOM literature, there has been an extensive
discussion of the form and existence of an objective function. Yet,
while the literature has provided a wide range of approaches for measuring
the quality of a SOM, there is still no commonly accepted objective
function for the general case (see, e.g., \citet{Yin2008}). This
has stimulated the use of a range of quality measures. The most common
measures are the standard quantization error $\varepsilon_{qe}$,
distortion measure $\varepsilon_{dm}$ \citep{LampinenOja1992} and
topographic error $\varepsilon_{te}$ \citep{Kiviluoto1996}, whereas
more recent work have extended these measures towards a range of directions.
Out of the mentioned measures, the two former ones quantify the quantization
quality of the map and the latter one the topological ordering of
the map. These have also been proposed as measures for the SOTM:

\begin{equation}
\varepsilon_{qe}=\frac{1}{T}\sum_{t=1}^{T}\frac{1}{N(t)}\sum_{j=1}^{N(t)}\left\Vert x_{j}(t)-m_{b(j)}(t)\right\Vert ,\label{eq:SOTM_QE}
\end{equation}

\begin{equation}
\varepsilon_{dm}=\frac{1}{T}\sum_{t=1}^{T}\frac{1}{N(t)}\frac{1}{M(t)}\sum_{j=1}^{N(t)}\sum_{i=1}^{M(t)}h_{ib(j)}(t)\left\Vert x_{j}(t)-m_{b(j)}(t)\right\Vert ,\label{eq:SOTM_DM}
\end{equation}

\begin{equation}
\varepsilon_{te}=\frac{1}{T}\sum_{t=1}^{T}\frac{1}{N(t)}\sum_{j=1}^{N(t)}u(x_{j}(t)),\label{eq:SOTM_TE}
\end{equation}
where $u(x_{j}(t))$ measures the average proportion of $x_{j}(t)\in\Omega(t)$
for which first and second BMU (within $A(t)$) are non-adjacent units. 

Yet, in order to provide one measure to steer the training of a SOM,
the focus herein is on a measure by \citet{KaskiLagus1996} that combines
quantization and topographic errors for evaluating the goodness of
a SOM. The Kaski-Lagus measure accounts for the distance between data
$x_{j}$ and BMU $m_{b}$ and the shortest path between the coordinates
of the first BMU $r_{b}$ and second BMU $r_{b'}$. Hence, it measures
both the accuracy in representing the data and the continuity of the
data-to-grid mapping. Formally, the goodness of the map is expressed
as follows:

{\scriptsize 
\begin{equation}
\varepsilon_{kl}=\frac{1}{T}\sum_{t=1}^{T}\frac{1}{N(t)}\sum_{j=1}^{N(t)}\left(\left\Vert x_{j}(t)-m_{b}(t)\right\Vert +\mbox{min}_{p}\sum_{g=1}^{P_{b'(x_{j}(t)),p}}\left\Vert m_{I_{p}(g-1)}(t)-m_{I_{p}(g)}(t)\right\Vert \right)\label{eq:SOTM_Kaski-Lagus}
\end{equation}
}where $b\left(x_{j}(t)\right)$ and $b'(x_{j}(t))$ represent the
index of the first BMU and second BMU and $I_{p}(g)$ denotes the
index of the path along the map grid from unit $I_{p}(0)=b(x_{j}(t))$
(BMU) to $I_{p}(P_{b'(x_{j}(t)),p})=b'(x_{j}(t))$. For $I_{p}$ to
be a path on the grid, the units $I_{p}(g-1)$ and $I_{p}(g)$ need
to be neighbors for $g=1,...,P_{b'(x_{j}(t)),p}$. Hence, the first
term is the standard quantization error, whereas the second term is
a continuity metric measuring whether close points in the mapped space
have contiguous BMUs on the map grid. Thus, when adapting the Kaski-Lagus
measure to each one-dimensional SOM of the SOTM, we have a single
criterion $\varepsilon_{kl}(t)$ to steer the choice of the parameter
$\sigma(t,s_{t})$ throughout the SOTM training. The implication of
a data-driven choice of $\sigma$  is not only an automation of the
SOTM training, but also the feature of adjusting the training to the
characteristics of the data at each time $t$. Likewise, in order
to derive qualities, or properties, of individual SOMs, we can compute
each of the quality measures $\varepsilon_{qe}$, $\varepsilon_{dm}$,
and $\varepsilon_{te}$ at the level of individual one-dimensional
SOMs, denoted $\varepsilon_{qe}(t)$, $\varepsilon_{dm}(t)$, and
$\varepsilon_{te}(t)$.

\subsection{Weighted SOTMs}

For a weighted approach, a SOM-based algorithm is suitable, as there
has been a number of works on weighted or cost-sensitive learning.
In the early days of the SOM, \citet{Kohonen1993} introduced weighting
of data in order to eliminate border effects on the edges of a SOM.
Similarly, in applications of the SOM to image compression, \citet{KimRa1993}
and \citet{Kangas1995} proposed the use of statistical properties
of data for weighting. Weighting can also be used to adjust for imbalanced
class frequencies and for varying the influence or priority of features
in SOM training (e.g., \citep{Vesantoetal1999}). The weighting in
the WSOM \citep{SarlinWSOM2012} is different as it provides means
for the user to specify the importance of data for learning, in the
unsupervised case for forming clusters. This is also the approach
that we propose to be used to weight the SOTM learning. 

Following \citep{SarlinWSOM2012}, the implemented weighted SOTM learning
is based upon the batch version of the SOM algorithm augmented with
an instance-specific weight. The weighted counterpart of the SOTM
iterates in two steps through $1,2\lyxmathsym{\ldots},t$, in a similar
fashion as the above presented original algorithm. The first step
follows the SOTM matching in Eq. \ref{eq:SOTM_match} by assigning
each input data vector $x_{j}(t)$ to its BMU $m_{c}(t)$. In the
second step, each prototype $m_{i}(t)$ (where $i=1,2\lyxmathsym{\ldots},M$)
is adjusted using a weighted counterpart of the batch update formula.
Hence, replacing Eq. \ref{eq:SOTM_update}, the update formula of
the weighted SOTM takes the following form: 

\begin{equation}
m_{i}(t,s_{t})=\frac{\sum_{j=1}^{N(t)}w_{j}(t)h_{ib(j)}(t,s_{t})x_{j}(t)}{\sum_{j=1}^{N(t)}w_{j}(t)h_{ib(j)}(t,s_{t})},\label{eq:WSOTM_update}
\end{equation}
where weight $w_{j}$ is the importance of $x_{j}(t)$ for the learning
of patterns. Moreover, as discussed in the beginning of this section,
but not further explored in the paper, other types of weighting is
obviously possible. For instance, for cases with prior knowledge on
the non-uniform importance of individual features, one might weight
or scale them to resemble differences in relevance.

\section{Experiments}

This section presents the schemes for automated and weighted SOTMs
on two real-world datasets: (\emph{i}) country-level risk indicators
to measure the evolution of global imbalances, and (\emph{ii}) credit
applicant data to measure the evolution of firm-level credit risks.

\subsection{From country risks to global imbalances}

Risk identification and assessment are two key tasks in financial
stability surveillance. Recently, in terms of country-level oversight
by authorities, the macroprudential approach has emerged as a key
means to safeguarding financial stability. Rather than only being
concerned with the failure of individual entities, a macroprudential
approach takes a holistic view on the financial system with the aim
and mandate to ensure system-wide stability. Over the past decades,
risk identification has mainly been attempted through predictive modeling,
which relies on historical data and conventional statistical methods
(see, e.g., \citep{Bergetal2005}) and computational intelligence
(see, e.g., \citep{SarlinPRL2013} and the review therein). Yet, given
the changing nature of financial crises, we have not been able to
avert from major periods of distress. Rather than mappings from high-dimensional
data to a crisis probability through predictive modeling, this motivates
exploratory analysis of the high-dimensional temporal and cross-sectional
data, not the least for exploring which risk and vulnerability indicators
are building up this time. Accordingly, \citep{SarlinPRL2013} proposed
a visual approach to identify the evolution of vulnerabilities and
risks through the standard SOTM. The application herein extends this
work by taking a truly macroprudential approach to the SOTM. While
the SOTM provides an illustration of the risks and vulnerabilities
in the cross-section, each entity has been assumed to be of equal
system-wide importance. The weighting scheme to the SOTM enables the
influence to be specified according to the systemic relevance of each
entity.

We apply a SOTM with an automated and weighted training scheme for
an abstraction of financial indicators before, during and after the
global financial crisis of 2007\textendash{}2009. As above mentioned,
the model herein improves the one in \citep{SarlinPRL2013}, and hence
the data used herein are also identical to the data in that work.
The dataset includes quarterly input matrices $\Omega(t)$, where
rows represent countries and columns binary class variables and 14
country-specific macro-financial indicators. The dataset consists
of 28 countries, 10 advanced and 18 emerging economies, from 1990:1\textendash{}2010:3.
The class variables represent pre-crisis, crisis, post-crisis and
tranquil periods and are objectively identified with a Financial Stress
Index (FSI). The FSI defines the crisis events, base upon which we
define the rest of the classes. The set of indicators are commonly
used measures in the macroprudential literature to capture build-ups
of domestic and global vulnerabilities and risks. More specifically,
they proxy credit developments and leverage, asset price developments
and valuations, as well as more traditional measures of macroeconomic
imbalances. Differences in indicators across countries are controlled
for by normalizing each input into historic country-specific percentiles.
To set the importance of each economy based upon their relevance to
the system, we define $w_{j}(t)$ to be the share of stock-market
capitalization of country $j$ in period $t$ of the sum of stock-market
capitalization in the sample in period $t$. While being an oversimplified
measure of an economy's systemic relevance, it is beyond the scope
of this paper to create a more advanced measure, which could be metrics
from network models, for instance.

\begin{figure}
\begin{centering}
\includegraphics[width=1\columnwidth]{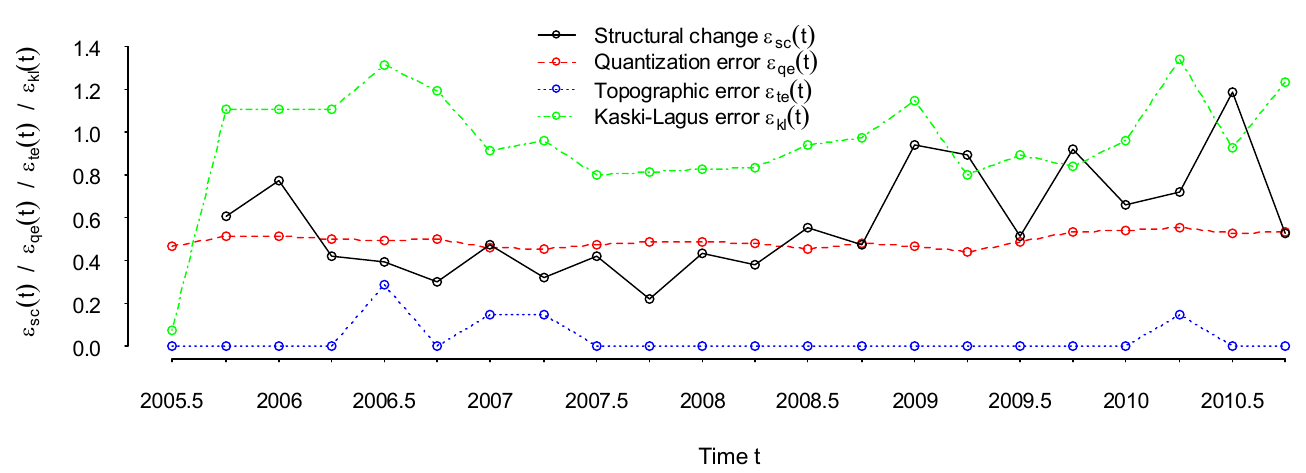}
\par\end{centering}

\textbf{\scriptsize Notes}{\scriptsize : The errors (}\textcolor{cyan}{\scriptsize $\varepsilon_{qe}(t)$}{\scriptsize ,}\textcolor{magenta}{\scriptsize{}
$\varepsilon_{te}(t)$}{\scriptsize{} and}\textcolor{green}{\scriptsize{}
$\varepsilon_{kl}(t)$}{\scriptsize ) are computed for time units
$t=1,2,\lyxmathsym{\ldots},T$ and }\textbf{\scriptsize $\varepsilon_{sc}(t)$}{\scriptsize{}
for time units $t=2,3,\lyxmathsym{\ldots},T$ for the final model
with an 7x22 array of units and a time-varying $\sigma(t)$.}{\scriptsize \par}

\centering{}\caption{\label{Fig:SOFSM_SOTM_property_measure} Quality and property measures
of the crisis SOTM. }
\end{figure}

The model architecture is set to 7x22 units, where 7 units represent
the cross-sectional dimension and 22 units the time dimension. While
the number of units along the quarterly time dimension is set as to
span periods before, during and after the crisis that started in 2007
(2005:2\textendash{}2010:3), the number of units depicting the cross-section
at each point in time is determined based upon its descriptive value.
Due to the property of a $A(t)$ approximating probability density
functions of the data $p(x,t)$, the SOTM, likewise the SOM, is not
restricted to treat each unit as an individual cluster. Hence, only
the dense locations in the data tend to attract units. As described
in Section 2, the Kaski-Lagus error measure $\varepsilon_{kl}(t)$
is used for choosing the final specification of the SOTM. For a SOTM
with 7x22 units, we let the neighborhood radius be a time-varying
parameter $\sigma(t)$ to optimize $\varepsilon_{kl}(t)$. Figure
\ref{Fig:SOFSM_SOTM_property_measure} shows how $\varepsilon_{kl}(t)$
varies over $t$. Increases in $\varepsilon_{kl}(t)$ can be seen
to correlate with increases in $\varepsilon_{qe}(t)$ and/or $\varepsilon_{te}(t)$.
An additional property metric measuring structural change (distance
between $A(t)$ and $A(t-\text{1})$) is denoted $\varepsilon_{sc}(t)$,
which illustrates that the largest changes in structures occur in
the latter part of the sample.

\begin{figure}
\begin{centering}
\includegraphics[width=0.9\columnwidth]{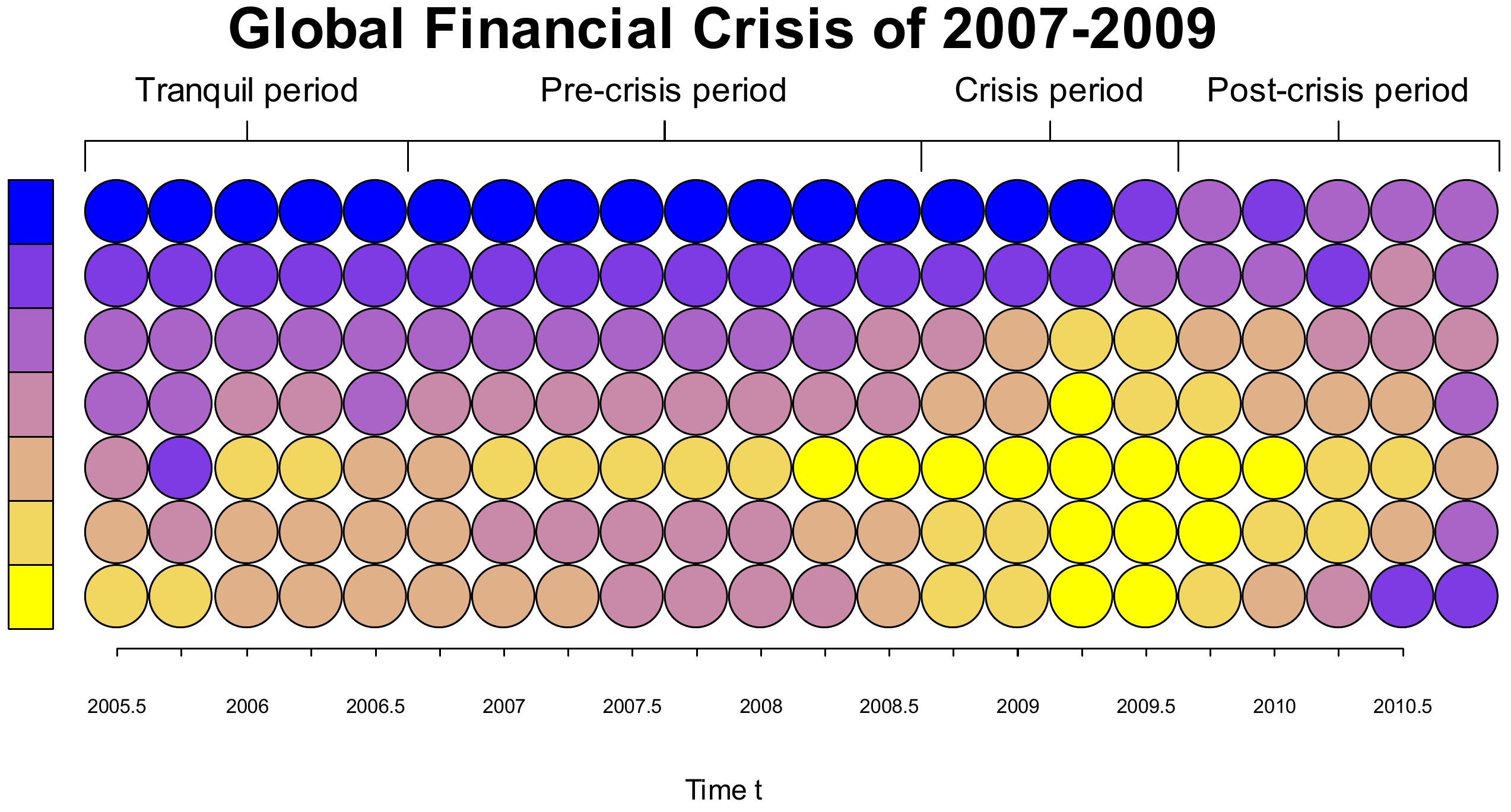}
\par\end{centering}

\textbf{\scriptsize Notes}{\scriptsize : The figure represents a SOTM
of the global financial crisis, where the coloring shows multivariate
differences in nodes. Labels above the figure define the classes in
data, i.e., the stages of the financial stability cycle.}{\scriptsize \par}

\centering{}\caption{\label{Fig:SOFSM_SOTM} A SOTM of the global financial crisis.}
\end{figure}

The SOTM is illustrated in Figure \ref{Fig:SOFSM_SOTM}, where the
labels above represent occurrences of the events in the cross-section
(averages of the class variables) and the timeline below the figure
represents the quarterly time dimension in data. The coloring of the
SOTM illustrates the proximity of units as approximated by Sammon\textquoteright{}s
mapping \citep{Sammon1969}. The differences of the units along the
vertical direction shows differences in cross-sections, while differences
along the horizontal direction shows differences over time. One can
observe that in the starting phases of the crisis in 2008, shifts
in the color scale towards yellow indicate a start of structural changes.
The structural changes reach their peak in 2009, while the structures
move back in mid-2010. First, we can observe that the crisis peak
is described by the shift towards yellow units. Second, we can observe
a post-crisis shift towards a lilac color. Comparing with the observed
patterns in \citep{SarlinPRL2013}, the shift towards values of distressed
economies (i.e., yellow) is larger when weighting according to stock-market
capitalization. This is also natural, as the largest countries were
also those suffering most from the crisis. 

\begin{figure}
\begin{centering}
\includegraphics[width=1\columnwidth]{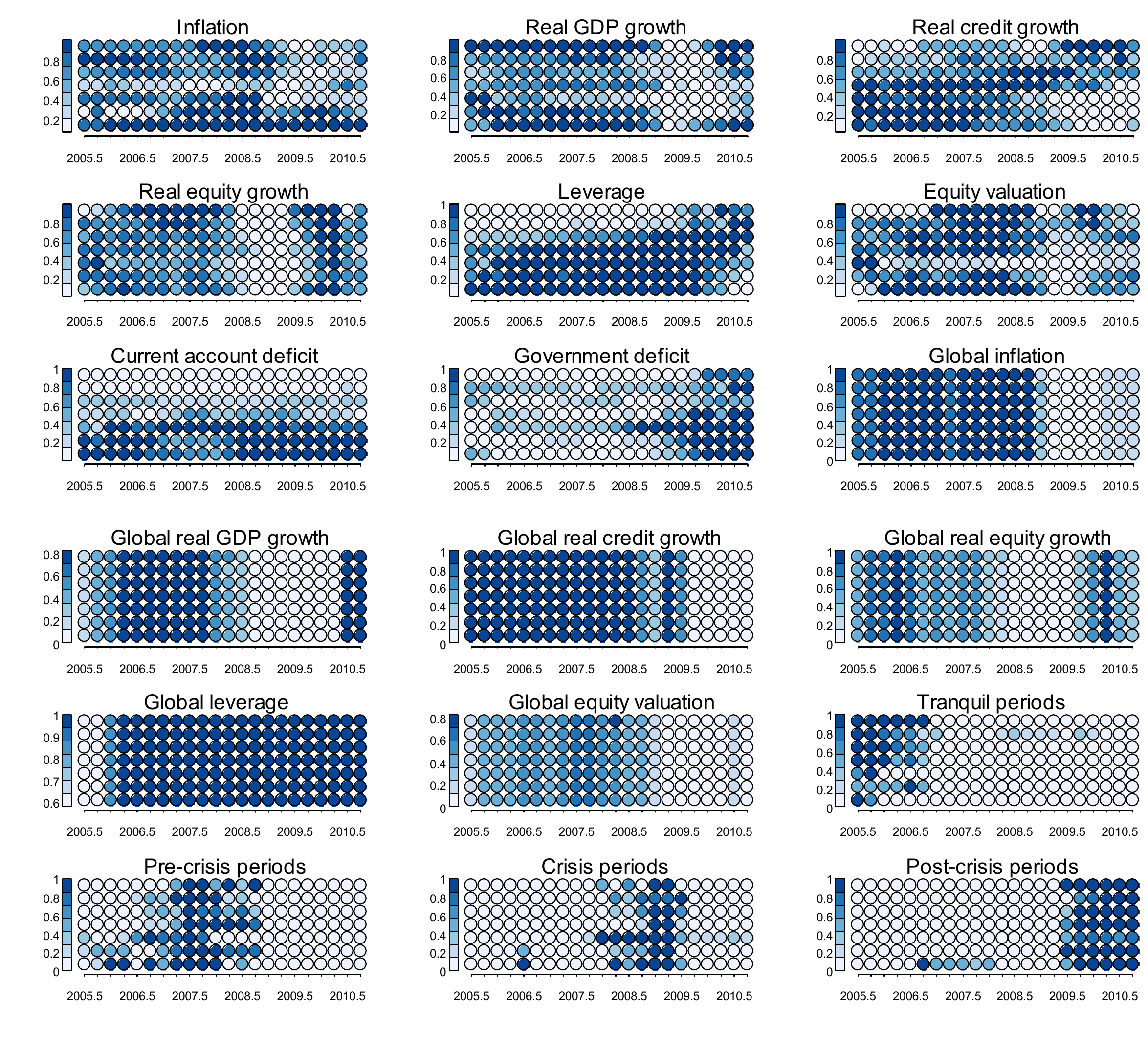}
\par\end{centering}

\textbf{\scriptsize Notes}{\scriptsize : The figure shows feature
planes for the 14 indicators and the class variables. The feature
planes are layers of the SOTM in Figure \ref{Fig:SOFSM_SOTM}. In
the case of binary class variables that take values 1 and 0, high
values represent a high proportion of data in that unit (pre-crisis,
crisis, post-crisis or tranquil periods).}{\scriptsize \par}

\centering{}\caption{\label{Fig:SOFSM_SOTM_FPs} Feature planes for the crisis SOTM. }
\end{figure}

For a more detailed view of individual risks and vulnerabilities,
we can explore the feature planes in Figure \ref{Fig:SOFSM_SOTM_FPs}.
They show for individual inputs (layers of Figure \ref{Fig:SOFSM_SOTM})
the spread of variable values using a blue scale (with individual
scales on the left, in which darker represents higher values). By
assessing the feature planes, one can visually both discover spread
of risks in the cross-section, and their variation over time. 

With no precise empirical treatment of differences between non-weighted
and weighted results, the feature planes in Figure \ref{Fig:SOFSM_SOTM_FPs}
generally illustrate expected differences. For instance, the post-crisis
period is depicted by a shift in countries being levered and exhibiting
high real credit growth, as well as countries with government deficits.
This illustrates that the global financial crisis was mainly driven
by imbalances in advanced economies, whereas the countries in the
rest of the world are exhibiting risks in credit development. As can
be expected, government deficits are shown to have widely increased
in the latter part of the analyzed period. In comparison to non-weighted
results in \citep{SarlinPRL2013}, we observe larger government deficits,
which also holds true with current increases in deficits in countries
with large financial systems, such as the United States and the euro
area.

\subsection{From credit applicants to firm-level risks}

In firms offering credit to customers, risk management focuses most
often on evaluating risks in individual customers. One concern in
the evaluation of credit risk is, however, the fact that credit scores
are non-linearly correlated. When risks build-up over time, and the
default probabilities increase, correlations in credit scores tend
to increase. Rather than only focusing on individual customers, we
take a holistic view on the credit risks building up in a company.
In particular, we make use of the SOTM to visualize how the customer
portfolio risks are building up over time. The weighting has a natural
interpretation also in this case. Each customer contributes to the
portfolio risk ($w_{j}(t)$) according to the amount of credit that
the firm has extended to the debtor at each time point $t$.

In this application, we apply a SOTM with an automated and weighted
training scheme for an abstraction of evolving risks in a company
providing credit to customers. The dataset used here consists of a
full loan database with all debtors for the firm over 4.5 years. The
dataset consists of close to 40,000 debtors with a total borrowing
of more than \$400 millions. Model architecture is set to 7x9 units,
where 7 units represent again the cross-sectional dimension, and 9
units represent the time dimension. The number of units along the
biannual time dimension is set as to span the available range of data,
whereas the number of units describing the cross-section is again
determined based upon its descriptive value. As in the previous application,
the neighborhood of the radius $\sigma(t)$ is optimized with respect
to $\varepsilon_{kl}(t)$.

\begin{figure}
\begin{centering}
\includegraphics[width=0.9\columnwidth]{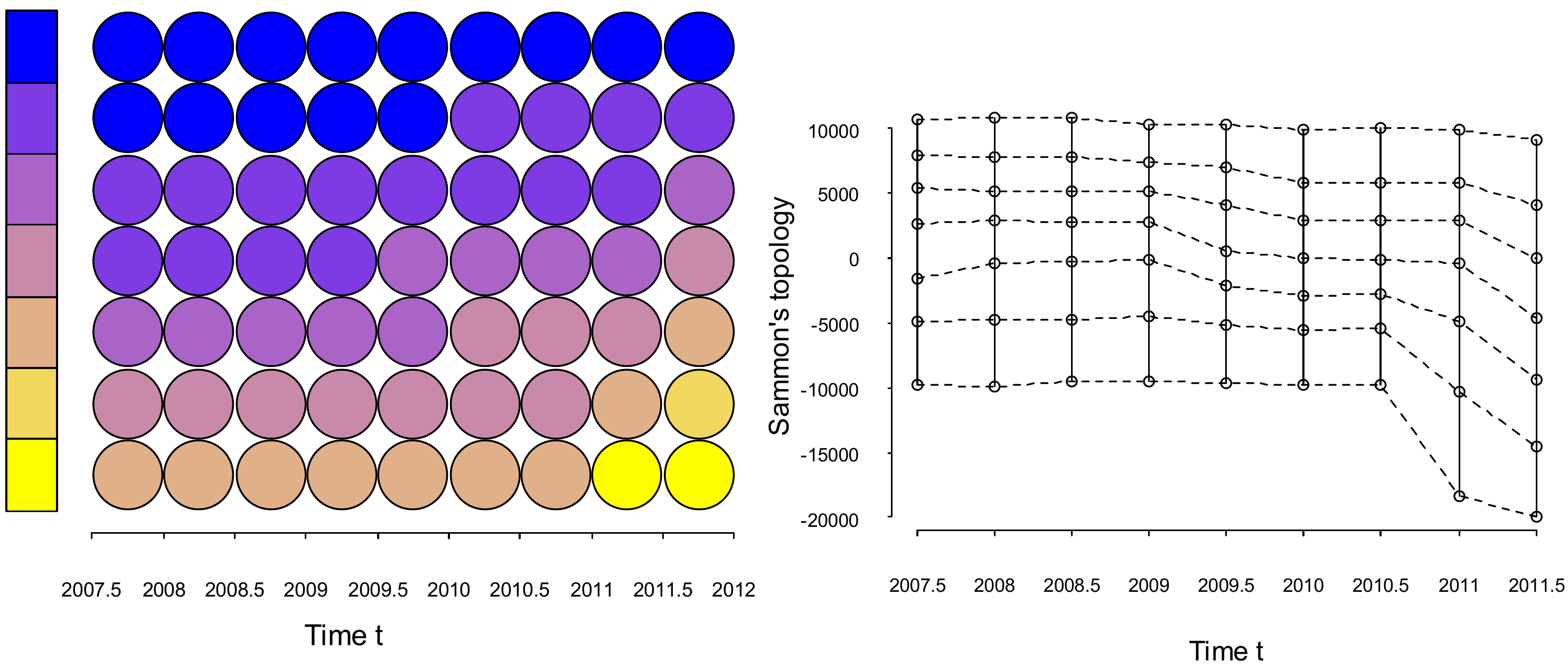}
\par\end{centering}

\textbf{\scriptsize Notes}{\scriptsize : The figure on the left represents
a SOTM of credit applicants in a firm, where the coloring shows multivariate
differences in nodes. The right figure shows the underlying Sammon's
topology.}{\scriptsize \par}

\centering{}\caption{\label{Fig:credit_SOTM} A SOTM of credit applicants in firm.}
\end{figure}

The credit SOTM is illustrated on the left in Figure \ref{Fig:credit_SOTM}.
As above, the coloring of the SOTM illustrates the proximity of units.
Further, the underlying Sammon's topology, which is used for the coloring,
is also shown on the right in Figure \ref{Fig:credit_SOTM}. The figures
mainly show that the upper part has been relatively stable, while
the lower part illustrates a new, emerging type of credit applicants.
The shifts in color on the left in Figure \ref{Fig:credit_SOTM} are
particularly well communicated with the line graph on the right.

\begin{figure}
\begin{centering}
\includegraphics[width=1\columnwidth]{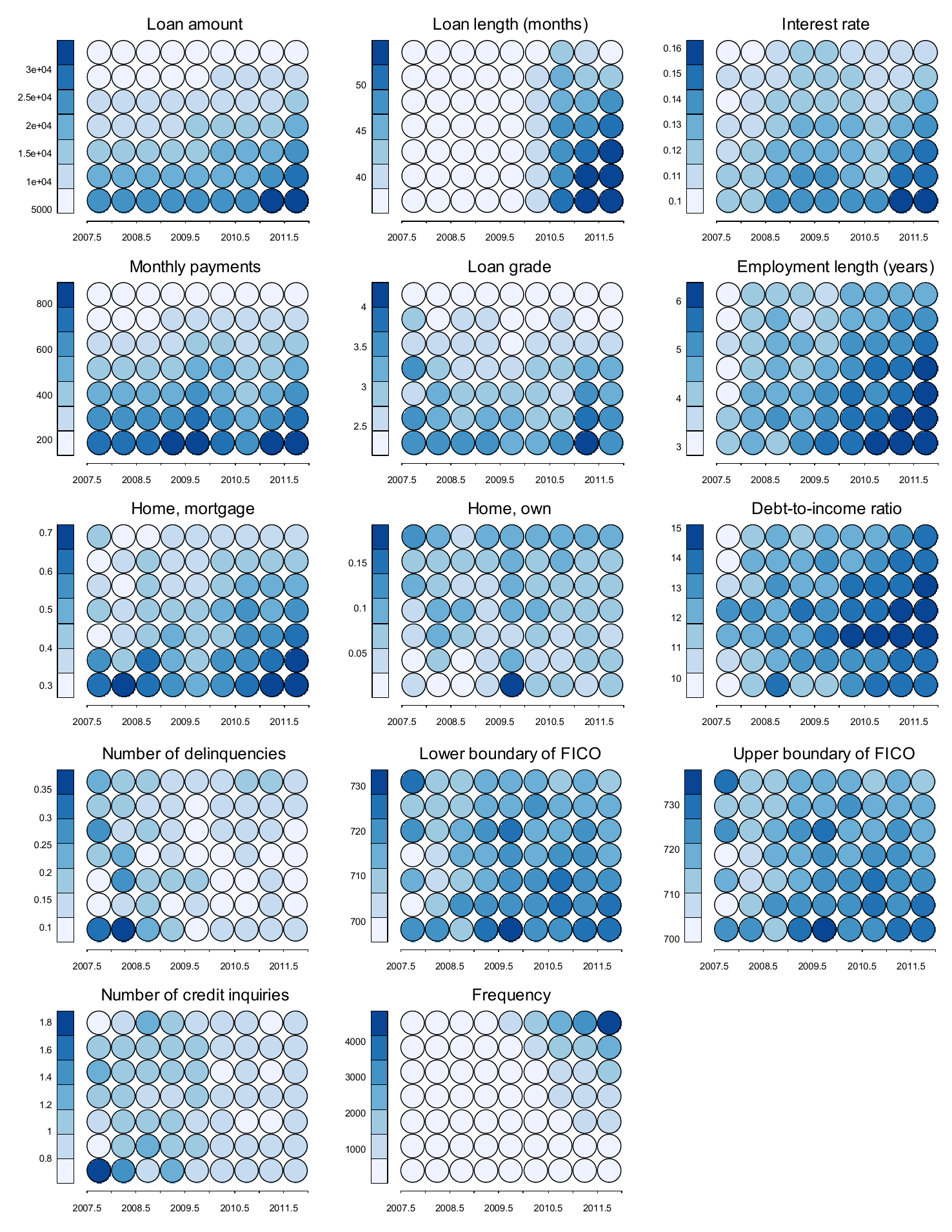}
\par\end{centering}

\textbf{\scriptsize Notes}{\scriptsize : The figure shows feature
planes for the 13 indicators and the frequency of data in each node.
The feature planes are layers of the SOTM in Figure \ref{Fig:credit_SOTM}.}{\scriptsize \par}

\centering{}\caption{\label{Fig:credit_SOTM_FPs} Feature planes for the credit SOTM. }
\end{figure}

To understand what drives the changes in credit applicants, we can
again view the feature planes. In Figure \ref{Fig:credit_SOTM_FPs},
the feature planes show the spread of individual variables on the
credit SOTM. Generally, with no focus on time, the lower part of the
map consists of customers with larger and longer loans, with higher
interest rates, lower grades and higher monthly payments. These customers
have more often a home mortgage, but luckily long historical employment
terms. When observing development over time, the feature planes illustrate
that loan lengths, sizes and interest rates have increased over time,
especially in the lower part of the map. Yet, when combining increases
in loans with the income of customers, we can observe that the debt-to-income
ratio has not increased most in the lower part of the map, but rather
in the center. Moreover, while the lower and central part of the map
can be assessed to comprise the customers of lowest grades, the largest
increases in frequency occurs in the upper part, which hence can be
assessed to indicate growth in healthy loans.

\section{Conclusions}

This paper has proposed two extensions to the SOTM: one for an automated
and one for a weighted training scheme. Motivated by the experience
in previous work on the SOTM, this paper aims at enhancing it to be
applicable in a wider setting. The rationale behind an automated SOTM
is not only a data-driven parametrization of the SOTM, but also the
feature of adjusting the training to the characteristics of the data
at each time step. The implication of a weighting scheme for the SOTM
is to improve learning from important data according to an instance-varying
weight. Here, importance might refer to a number of notions, such
as the certainty or trustworthiness of a data point or the relative
relevance of an instance. The schemes for automated and weighted SOTMs
have been illustrated on two real-world datasets: (\emph{i}) country-level
risk indicators to measure the evolution of global imbalances, and
(\emph{ii}) credit applicant data to measure the evolution of firm-level
credit risks. Among the applications of the SOTM mentioned in the
introduction, the weighting scheme would be suitable for illustrating
how green customer segments are evolving over time, where the greenness
degree, as estimated from buying behavior, would be the user-specified
weight. The general nature of the automation would, on the other hand,
be suitable for most cases.

\section*{References}

\bibliographystyle{elsarticle-harv}
\addcontentsline{toc}{section}{\refname}\bibliography{References/references}

\end{document}